\documentclass[final,3p,times,twocolumn,authoryear]{elsarticle}

\usepackage{amssymb}
\usepackage{amsmath}
\usepackage{graphicx}    
\usepackage{subcaption} 
\usepackage{tabularray}
\usepackage{xcolor}

\usepackage{hyperref}

\begin{document}

\begin{frontmatter}





\title{AGRICAM: A Track-Mounted Crop Pollination Monitoring Robot} 


\author[a]{Malika Nisal Ratnayake\corref{cor1}}
\ead{malika.ratnayake@monash.edu}

\author[b]{Adel N. Toosi}
\ead{adel.toosi@unimelb.edu.au}

\author[c]{James Cook}
\ead{james.cook@westernsydney.edu.au}

\author[d]{Romina Rader}
\ead{rrader@une.edu.au}

\author[a]{Alan Dorin}
\ead{alan.dorin@monash.edu}

\cortext[cor1]{Corresponding author.}

\affiliation[a]{organization={Department of Data Science and Artificial Intelligence, Monash University},
            addressline={Wellington Road},
            city={Clayton},
            state={VIC},
            postcode={3800},
            country={Australia}}

\affiliation[b]{organization={School of Computing and Information Systems, The University of Melbourne},
            addressline={Parkville},
            city={Melbourne},
            state={VIC},
            postcode={3052},
            country={Australia}}

\affiliation[c]{organization={Hawkesbury Institute for the Environment, Western Sydney University},
            addressline={Locked Bag 1797},
            city={Penrith},
            state={NSW},
            postcode={2751},
            country={Australia}}

\affiliation[d]{organization={School of Environmental and Rural Science, University of New England},
            city={Armidale},
            state={NSW},
            postcode={2351},
            country={Australia}}
            
\begin{abstract}
Insect pollination is critical for global food production, yet monitoring pollinators at commercial farm scale remains a challenge. Recent advances in computer vision and deep learning have enabled detailed analysis of pollinator behaviour, but monitoring must trade-off detail against spatial coverage and human or technological resources. {This paper presents the Automated Guided Robot for Insect and Crop Activity Monitoring (AGRICAM), a purpose-built robotic system designed to meet the requirements of large-scale pollination monitoring in protected cropping systems.} AGRICAM operates autonomously on low-cost, easily installed track for movement along crop rows, without disrupting farm operations or insect behaviour. The platform integrates two RGB cameras, microclimate sensors, GPS and RFID modules, motion sensors, and 4G cellular network connectivity for data transmission. A web interface enables remote device configuration and scheduling. The system autonomously captures video and image data of insects' locations and local environmental conditions. These are transferred to the cloud and analysed using computer vision models to quantify pollinator visitation and spatio-temporal activity variation. We deployed the system on a commercial blueberry farm to demonstrate and test its capability. It successfully mapped insect pollination patterns across 80 m long industrial polytunnels over 30 hours. This data enabled spatial analyses of insect activity we used to confirm a uniform pollinator distribution within polytunnels, as desired by the farm management team. The data also highlighted variation of insect activity associated with time of day and microclimate. AGRICAM therefore has been shown to be a scalable, automated crop pollination monitor that can support data-driven decisions to enhance pollination management, thereby improving crop productivity and food security.

\end{abstract}



\begin{keyword}

Mobile robotics; Precision agriculture; Pollination monitoring; Computer vision; Artificial intelligence



\end{keyword}

\end{frontmatter}



\section{Introduction}
\label{sec:introduction}

Pollinators play a key role in food production and ecosystem management. Animal-mediated pollination contributes to approximately 35\% of global crop production volume~\citep{fao2018} and supports nearly 75\% of the world’s leading food crops, including more than 87 high-value crops directly consumed by humans~\citep{potts2016assessment, klein2007importance}. In addition to enhancing crop quantity, pollination services improve crop quality~\citep{tscharntke2025pollinator}, nutritional value~\citep{ellis2015pollinators}, and diversity~\citep{shrestha2008honeybees}. The global economic value of pollination services has been estimated between USD 235B and 577B annually, highlighting their substantial contribution to agricultural economies ~\citep{potts2016assessment}.

Modern agriculture relies on the honeybee, \textit{Apis mellifera}, as the primary managed pollinator because it is easy to manage~\citep{potts2010global} and can pollinate a wide range of crops. However, wild pollinators, including bumblebees, solitary bees, flies, moths, butterflies, and other insects, also contribute significantly to crop productivity and often enhance yield stability and fruit quality~\citep{garibaldi2013wild, macinnis2019pollination, rader2016non}. In recent decades, key wild and managed pollinator populations have suffered under interacting stressors such as climate change~\citep{vasiliev2021role, bonebrake2012climate}, habitat loss and fragmentation~\citep{kline2020mitigating}, urbanisation~\citep{sanetra2024disentangling}, pesticide exposure~\citep{brittain2010impacts}, and emerging parasites (e.g. Varroa destructor)~\citep{wilfert2016deformed}. Larger insect populations are desirable as they are frequently associated with higher crop yields~\citep{rollin2019impacts}. However, the effectiveness of different pollinator populations is also important, and differs between crops~\citep{macinnis2019pollination}. Therefore, improved understanding and management of pollinators through timely and efficient monitoring of pollinator–crop interactions is essential to understand and sustain crop productivity and ensure long-term pollinator health~\citep{garibaldi2020crop}. This study is motivated by this need and aims to develop automated approaches for pollination monitoring and management.

Monitoring insect pollinators is essential for understanding their spatial distribution within agricultural landscapes and quantifying the contribution of different species to crop pollination. Conventional monitoring approaches include transect walks, direct flower observations and passive sampling~\citep{howard2021towards}. While these methods may be relatively straightforward to implement, they are also time-consuming, labour-intensive and require expertise to identify species concerned. Additionally, reliance on human observers may introduce unintentional bias, reduce reproducibility, and increase processing time and operational costs~\citep{dennis2006effects, simons1999gorillas}. Passive insect sampling methods, including pan traps and other static trapping techniques, reduce labour requirements at the time of sampling but require (usually manual) processing at a later stage to interpret data. Further, passive sampling does not record actual flower visitation events and therefore provides limited insights into species-specific contributions to pollination. Consequently, while passive methods assist in determining pollinator abundance, they do not capture the interactions required to understand pollinator effectiveness.

Recent advances in sensing technology and artificial intelligence have enabled the development of automated pollinator monitoring systems. Some approaches infer pollination activity by monitoring hive-level indicators such as hive entrance traffic~\citep{babic2016pollen,ngo2021automated}. However, because these measures don't specifically document crop foraging, they may mask inadequate pollination, for instance, if honeybees preferentially forage on non-target vegetation~\citep{howard2021towards}. Other commercial systems use distributed microphones and machine learning to detect wingbeat frequencies and estimate pollinator abundance~\citep{crochard2026buzzy, hearon2025buzzdetect}. While energy-efficient and relatively cheap, these methods do not directly measure flower visitation. Hence, the data they collect may be difficult (or impossible) to convert to an understanding of pollination when monitoring unfamiliar insects whose crop visitation patterns are poorly understood.

Computer vision and deep learning have increasingly been applied in agriculture for tasks such as fruit detection~\citep{afonso2020tomato}, yield estimation~\citep{koirala2019deep}, weed detection~\citep{su2021data}, and both pest and beneficial insect monitoring~\citep{amarathunga2021methods, amarathunga2022fine}. More recently, these techniques have been used to monitor pollination by deploying cameras in crops to detect, classify, and track insects. However, automated monitoring in agricultural environments remains challenging due to dynamic conditions including fluctuating illumination, wind-driven plant movement, background clutter, and everyday farm operations. These factors complicate reliable detection of small, fast-moving, unmarked pollinators. Some studies have reduced this complexity by attracting insects to artificial uniform backgrounds \citep{sittinger2024insect}, but such approaches do not record crop flower visits essential for quantifying pollination activity.

Several studies have implemented computer vision systems to directly observe insects on crop flowers~\citep{ratnayake2021towards, ratnayake2023spatial}, providing functionally relevant, high-resolution temporal pollination data. However, these systems typically rely on fixed cameras with a field of view limited to a small number of flower clusters \citep{ratnayake2023spatial}. In large protected environments such as polytunnels (50 - 100 meters in length), this spatial coverage is inadequate as pollinator activity varies with microclimatic variations in temperature, humidity, light, and airflow \citep{hall2020bee}. Some developers have used drone-technologies to attempt to expand spatial monitoring by estimating pollination indirectly through floral cover and tracking tagged insects~\citep{torresani2023novel, shearwood2020c}. However, these systems do not directly measure pollination, are of high complexity, and potentially disturb the pollinator behaviour they are trying to monitor~\citep{batsleer2020neglected}. 

Agricultural robotics, increasingly used for monitoring, weeding, fertilising and harvesting, offer a promising alternative~\citep{botta2022review}. Robotic pollination systems have been explored for crops requiring manual intervention due to pollinator scarcity or specialised pollination mechanisms, such as buzz pollination~\citep{gao2023novel, singh2025comprehensive}. However, replacing natural pollinators is impractical and unsustainable, as robots cannot replicate their ecological complexity or ecosystem services~\citep{nimmo2022replacing, potts2010global, gleadow2019averting}. Instead, mobile robotic platforms can better complement natural systems by automating and expanding spatiotemporal pollination monitoring coverage to detect localised deficits and support adaptive management.

Therefore, there is a clear need for scalable monitoring systems that capture fine resolution spatial variation in pollinator activity across large protected cropping environments. Ideally this technology would enable continuous, automated assessment of flower visitation patterns across heterogeneous microclimates within cropping infrastructure. Understanding such insect spatial dynamics is essential for optimising pollinator management, improving crop yield, and improving overall sustainability of cropping systems~\citep{sritongchuay2026crop}. Here, we present a novel mobile robotic system designed to meet all of these needs through the following contributions:

\begin{enumerate}
    \item Design and development of AGRICAM, a suspended track-based mobile robotic system for autonomous pollination monitoring in protected cropping. 
    \item Integration of vision-based and microclimate sensing, RFID-based localisation, 4G connectivity, and a web interface for real-time data acquisition and remote control. 
    \item {Implementation of a computer vision pipeline for high-resolution spatiotemporal analysis of pollinator activity in three-dimensional crop canopies.} 
    \item Field validation in a commercial blueberry farm demonstrating spatiotemporal mapping of pollinator activity and its relationship with microclimate.
\end{enumerate}

The remainder of the paper is organised as follows. Section~\ref{sec:methods} describes the system design and implementation, Section~\ref{sec:results} presents field evaluation and pollination monitoring outcomes, and Section~\ref{sec:discussion} discusses the implications, limitations, and future research directions. Section~\ref{sec:conclusion} concludes the paper.

\section{Methodology}
\label{sec:methods}
\begin{figure*}[h]
    \centering
    \includegraphics[width=\textwidth]{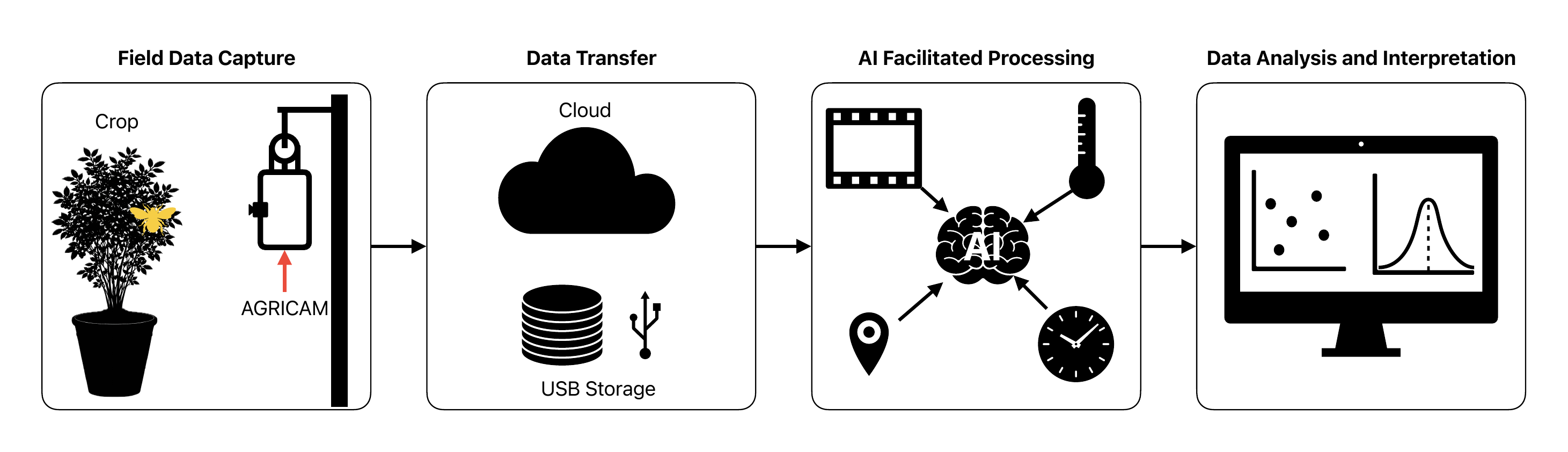} 
    \caption{\textbf{Overview of the AGRICAM processing pipeline.} Video, location, and microclimate data are collected in the field using AGRICAM, transferred via cloud (or USB storage if desired), and processed by AI-facilitated workflows to automatically detect and track insect pollinators. The data are subsequently analysed and visualisations of spatiotemporal patterns of microclimate and insect behaviour are prepared.}
    \label{fig:agricam_overview}
\end{figure*}

Figure~\ref{fig:agricam_overview} presents an overview of the pollination monitoring system and its data processing pipeline from field data acquisition and data transfer, through AI-facilitated processing, to data analysis and interpretation.

\subsection{System Design}

\begin{figure*}[h!] 
    \centering 

    \begin{subfigure}[t]{0.98\textwidth}
        \centering
        \includegraphics[width=\textwidth]{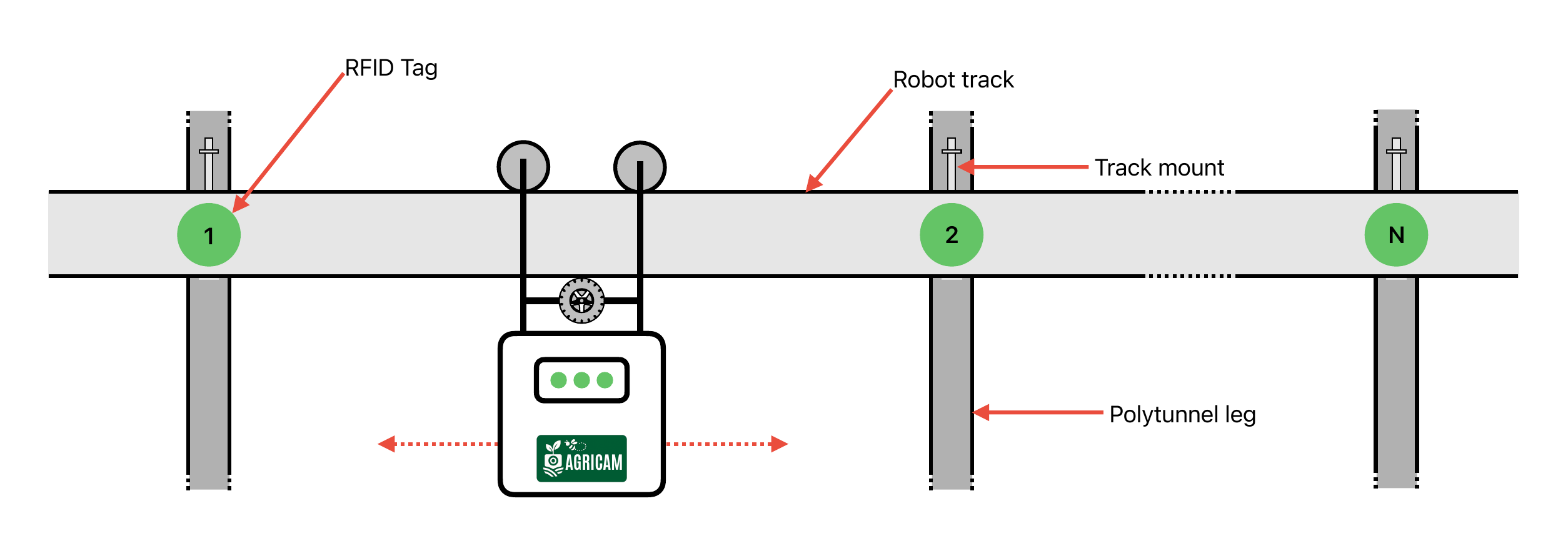}
        \caption{}
        \label{fig:system_installation_a}
    \end{subfigure}
    
    \vspace{1em} 

    \begin{subfigure}[t]{0.60\textwidth}
        \centering
        \includegraphics[width=\textwidth]{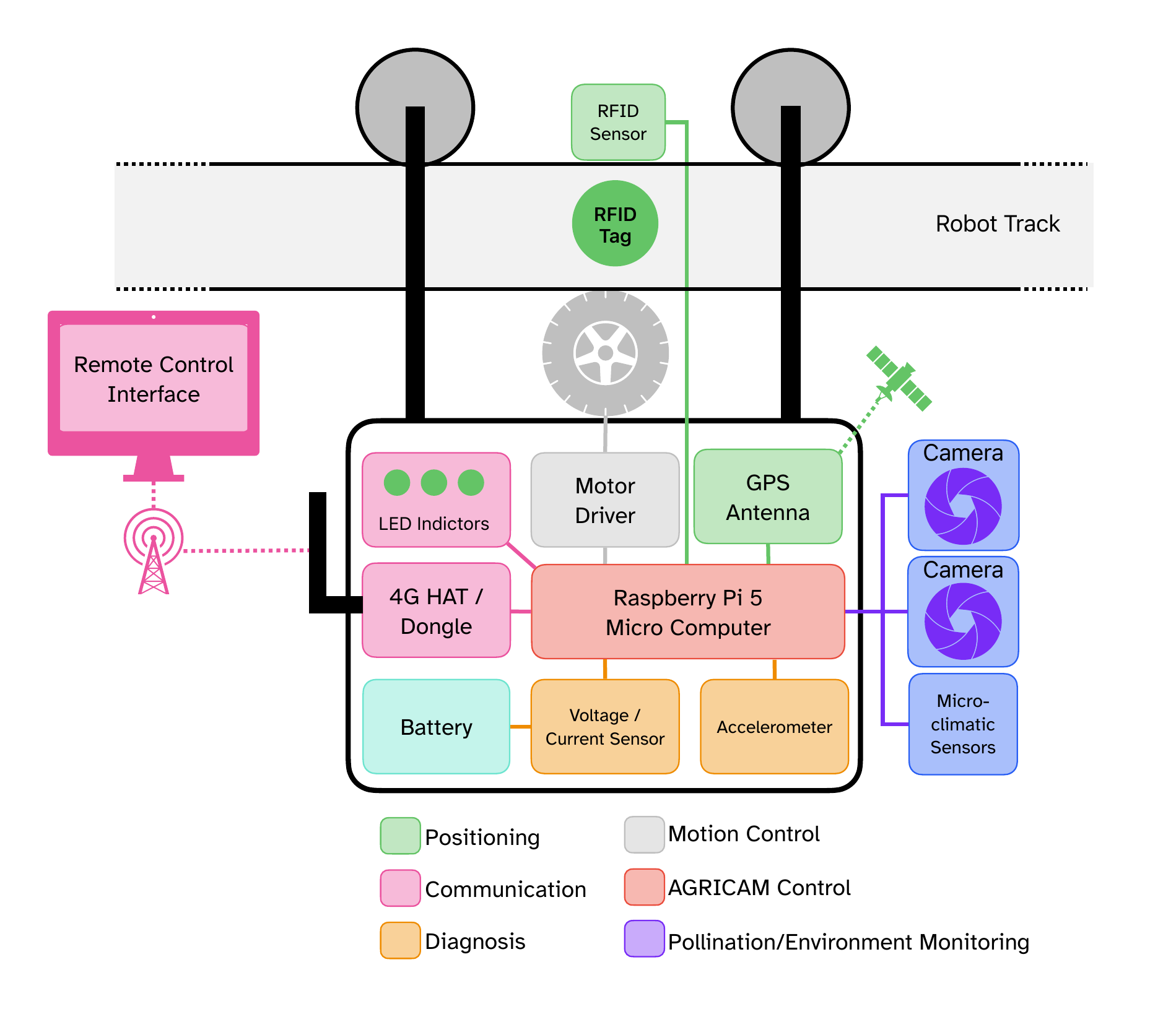}
        \caption{}
        \label{fig:system_installation_b}
    \end{subfigure}
    \hfill
    \begin{subfigure}[t]{0.36\textwidth}
        \centering
        \includegraphics[width=\textwidth]{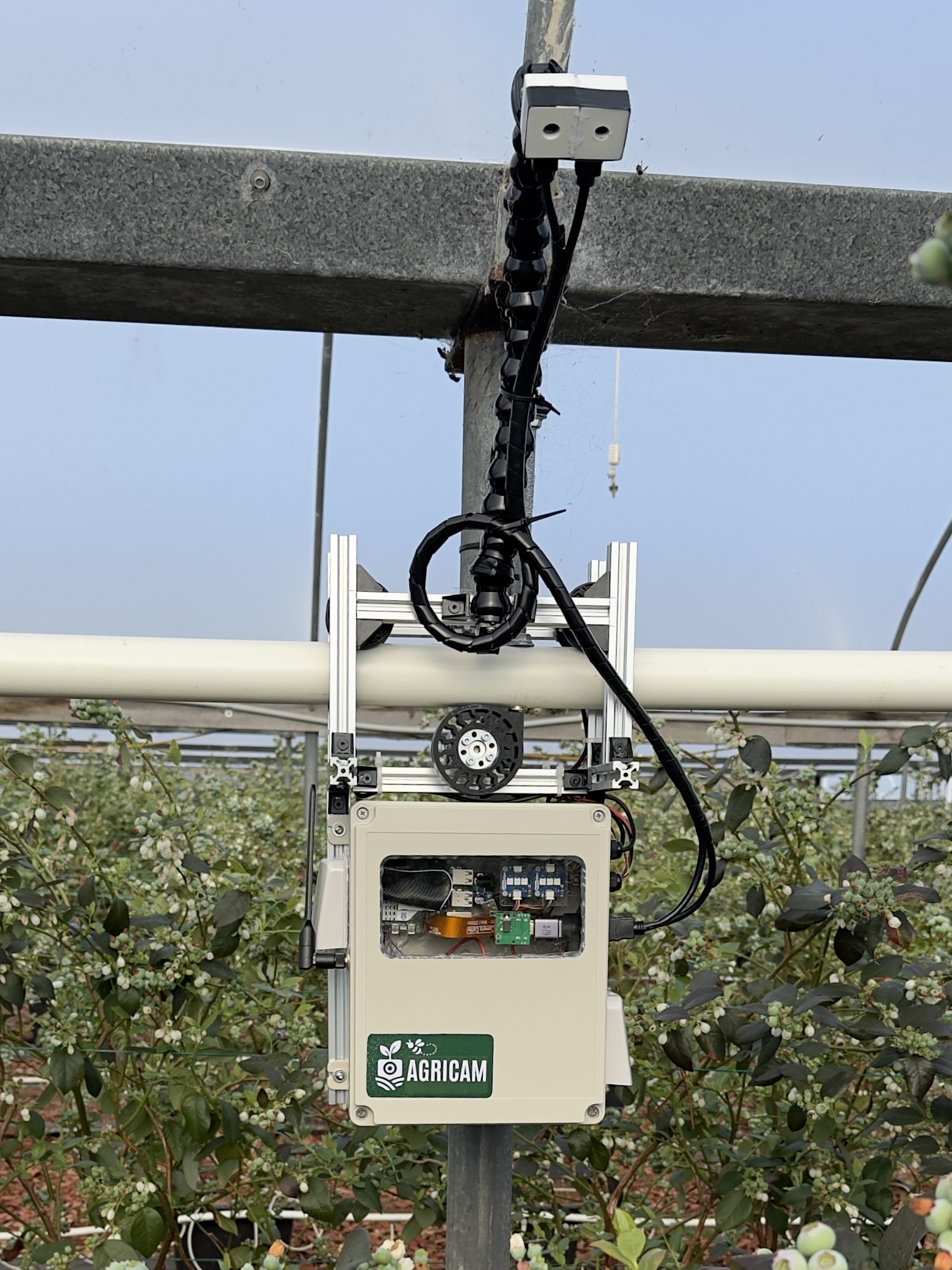}
        \caption{}
        \label{fig:system_installation_c}
    \end{subfigure}

    \caption{\textbf{AGRICAM system components}. (a) System installation overview, (b) components of the AGRICAM robot, and (c) deployed system in a commercial blueberry farm. In (a) and (c), the AGRICAM unit is mounted on a PVC track along the leg rows of a protected cropping structure (polytunnel). Each monitoring station is identified by an RFID tag and positioned at structurally supported locations to minimise stress on the PVC track. In (b), task-specific components of the AGRICAM system are shown in different colours.}
    \label{fig:system_installation_overview}
\end{figure*}

AGRICAM (Figure \ref{fig:system_installation_overview}) was developed to operate autonomously and reliably within protected cropping environments. Key hardware design goals included robustness to variable environmental conditions, and to minimise human intervention, disruption to pollinators and farm operations. To satisfy these requirements, the system is housed in a water-resistant enclosure and rolls along a suspended track. This configuration eliminates ground contact to prevent soil compaction, wheel slippage, and obstruction of harvesting pathways, while ensuring a consistent travel path and repeatable camera positioning within space-constrained protected cropping infrastructure.

\subsubsection{Robot track}

The low maintenance track is installed inside a polytunnel adjacent to crop rows to provide a stable platform for robot mobility while minimising interference with routine agricultural machinery and farm activities. The track is fabricated using conventional agricultural polyvinyl chloride (PVC) pipes suspended from polytunnel legs using angle brackets and threaded rods. PVC was selected as it is cheap, light, easy to work with and a basic consumable found on farms. The threaded rods facilitate track height and inclination adjustment to match the crop's flowering height.

\subsubsection{Navigation and positioning}
Accurate positioning of the AGRICAM robot within the crop is essential to ensure precise, repeatable data collection. Previous research has employed Global Navigation Satellite Systems (GNSS)~\citep{leanza2023heading, galati2022robonav} and vision-based odometry~\citep{ball2016vision, han2023visual} for localisation in agricultural robotics. However, GNSS-based positioning can sometimes be unreliable  under protective structures due to signal attenuation and reflections caused by metallic frames and coverings. Moreover, the typical positioning accuracy of standard GNSS systems (~1 m) is inadequate for our application.

Although vision-based odometry offers robustness under varying environmental conditions, it is highly sensitive to illumination changes and performs poorly in polytunnels without the addition of landmarks. To overcome these limitations, Radio Frequency Identification (RFID) localisation was implemented using pre-programmed RFID tags installed along the track, each encoded with a unique station identifier, and a robot-mounted RFID reader. This enables the robot to determine its position with precision, independent of environmental lighting or weather conditions. During navigation, the RFID reader continuously scans for nearby tags. Upon detecting a tag, the robot initiates a deceleration sequence and stops to confirm the tag’s identification number. This two-phase reading strategy minimises the likelihood of misreading tags during high-speed movement.

In cases where a tag read fails, the robot estimates the current station number based on its previous position, direction of movement, and the total number of stations along the track. The estimation process is expressed as:
\begin{equation}
S_{\text{est}} =
\begin{cases}
\min(S_{\text{prev}} + 1, S_{\max}), & \text{if } D = \text{toward } S_{\max} \\
\max(S_{\text{prev}} - 1, S_{\min}), & \text{if } D = \text{toward } S_{\min}
\end{cases}
\end{equation}

\noindent where $S_{\text{est}}$ is the estimated station number, $S_{\text{prev}}$ is the previously recorded station, $S_{\max}$ and $S_{\min}$ denote the upper and lower limits of the track, respectively, and $D$ represents the direction of robot movement. When the robot moves toward $S_{\max}$, the estimated station index increments by one until it reaches the upper limit, whereas motion toward $S_{\min}$ decrements the index bounded by the lower limit. 

To complement RFID-based localisation, a GNSS module is integrated into the robot to obtain global position data when available, enabling the system to record both local (relative) and global (absolute) positions. Additionally, limit switches are installed at the extreme ends of the track as a safety mechanism. If the RFID reader fails to detect tags, the robot continues to move until it activates a limit switch, upon which it automatically reverses its direction of travel.

\subsubsection{Path planning and scheduling}

The robot’s path is calibrated based on the spatial positions of RFID tags installed along the track. The stopping stations at which the robot halts are fully programmable, allowing users to define a subset of tags as monitoring stations according to the spatial resolution required for pollination data collection. At each station, the robot captures video footage and environmental sensor data (Sec.~\ref{subsec:monitoring}). A designated \textit{home station} serves as the start and end point for each monitoring run. A complete traversal of all assigned stopping stations, beginning and ending at the home station, is defined as a single \textit{monitoring cycle}. The duration between consecutive cycles is user-defined to acquire data at the desired temporal resolution.

At the start of each day, a monitoring schedule is automatically generated based on the user-specified start time, end time, and monitoring cycle interval. If the robot fails to return to the home station before the start of the next scheduled cycle, that cycle is skipped, and the robot resumes operation at the subsequent scheduled time. At the end of the day, the robot returns to a predefined \textit{safe station} to minimise the risk of vandalism, theft, or environmental exposure during inactivity.

\subsubsection{Motion control}
The AGRICAM robot is suspended from the track using four idler wheels and driven by a single high-grip flexible wheel positioned beneath the track. The drive wheel has compliant spokes that provide traction and allow the robot to traverse minor variations in the track surface while maintaining stability and reducing mechanical stress. The drive wheel is powered by a 12 V direct-current (DC) gear-motor with a 70:1 reduction ratio, selected to provide sufficient torque to overcome the typical inclinations (< 10 degrees) observed within the protected cropping environment. This single-motor configuration keeps energy consumption of the robot low to extend battery life.

Motor control is achieved using a high-power dual-channel motor driver, interfaced with a Raspberry Pi microcontroller control unit and powered by a 12 V battery. Under standard operating conditions, the motor operates at approximately 25\% of its maximum power capacity. Pulse-width modulation (PWM) signals are used to regulate motor speed for gradual deceleration and acceleration at stopping stations.

\subsubsection{Communication and user interface}

AGRICAM incorporates 4G network connectivity to enable remote monitoring and control via a web interface. The interface is hosted locally on the robot, with access granted to authorised users via a secure tunnelling protocol. This configuration allows real-time interaction with the robot without the need for a dedicated external server. 4G was selected over alternatives such as Wi-Fi or LoRa, as it enables straightforward remote connectivity via a SIM card, providing wide-area coverage and sufficient bandwidth for reliable video data transmission and real-time control~\citep{klimiashvili2020lora}. 

\begin{figure*}[h!] 
    \centering 

    \begin{subfigure}[t]{0.48\textwidth}
        \centering
        \fbox{\includegraphics[width=\textwidth]{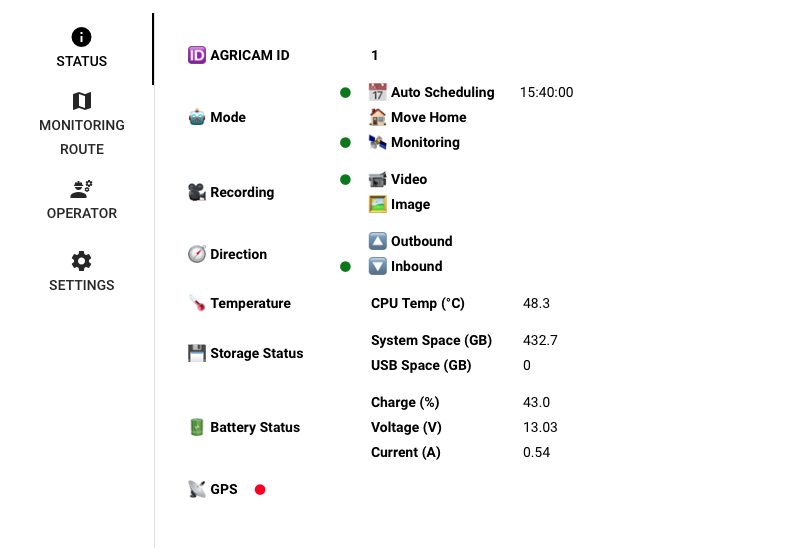}}
        \caption{}
        \label{fig:interface_home}
    \end{subfigure}
    \hfill
    \begin{subfigure}[t]{0.48\textwidth}
        \centering
        \fbox{\includegraphics[width=\textwidth]{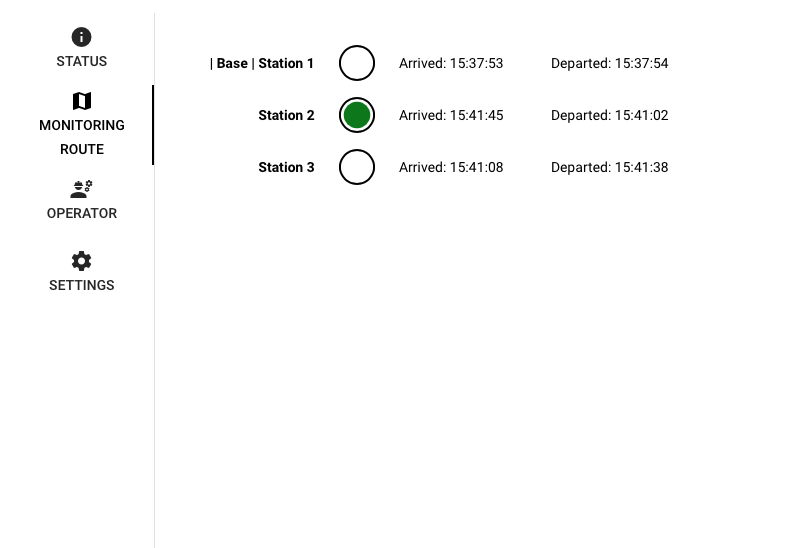}}
        \caption{}
        \label{fig:interface_map}
    \end{subfigure}

    \vspace{1em} 

    \begin{subfigure}[t]{0.48\textwidth}
        \centering
        \fbox{\includegraphics[width=\textwidth]{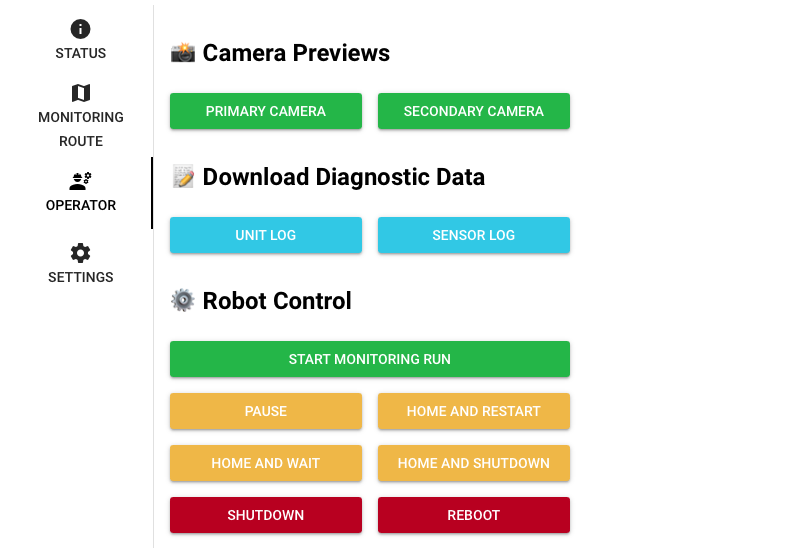}}
        \caption{}
        \label{fig:interface_control}
    \end{subfigure}
    \hfill
    \begin{subfigure}[t]{0.48\textwidth}
        \centering
        \fbox{\includegraphics[width=\textwidth]{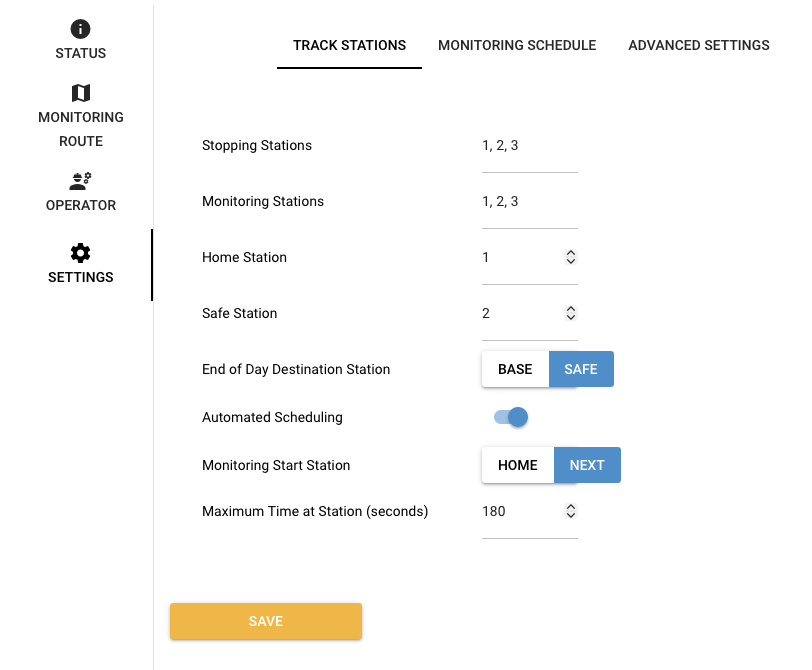}}
        \caption{}
        \label{fig:interface_settings}
    \end{subfigure}

    \caption{AGRICAM web user interface. (a) System status dashboard, (b) monitoring path map, (c) control and diagnostics panel, and (d) configuration menu.}
    \label{fig:system_interface}
\end{figure*}

The web control interface has four primary modules (Figure~\ref{fig:system_interface}):
\begin{enumerate}
    \item {System status dashboard} – provides an overview of the robot’s operational state, battery level, and sensor health.
    \item {Monitoring path map} – displays the robot’s movement path and logs arrival and departure times at each monitoring station.
    \item {Control and diagnostics panel} – offers live access to the robot camera view and options to download sensor data, diagnostic logs, and manually control robot motion.
    \item {Configuration menu} – enables users to modify robot parameters and initiate setup procedures.
\end{enumerate}

In addition to the web interface, an LED array visible through the robot housing conveys connectivity status, power warnings, and system faults for in-field diagnostics.

\subsubsection{Power Management}
AGRICAM is powered by a 12~V, 12~Ah lithium iron phosphate (LiFePO$_4$) battery selected for its high energy density, light weight, and enhanced safety compared with lead–acid or lithium-ion batteries.

Power from the battery was interfaced through a voltage, current, and power sensing module to monitor charge level and energy consumption before distribution to the control and actuation subsystems. The supply was divided in parallel between the Raspberry~Pi microcomputer and the motor driver unit. The connection to the Raspberry~Pi was regulated to 5~V using a step-down (buck) DC–DC converter. The connection to the motor driver was routed through a relay-controlled switch, allowing the Raspberry~Pi to completely disconnect power to the drive system when the robot is idle.


The voltage sensor continuously monitors the battery voltage, which is used to estimate the state of charge (SoC) as a percentage. The measured voltage is linearly interpolated between predefined minimum and maximum voltage thresholds corresponding to 0\% and 100\% charge, respectively.  When the estimated SoC falls below a predefined threshold, the robot autonomously returns to the base station and powers down to prevent over-discharge.

\subsubsection{Pollination and environmental monitoring}
\label{subsec:monitoring}

AGRICAM is equipped with two Raspberry Pi Camera Module V3 units, each incorporating a 12-megapixel Sony IMX708 CMOS image sensor with an autofocus lens for monitoring insect pollinators. The inclusion of two cameras increases spatial coverage. The cameras can be configured to capture time-lapse and/or video data either simultaneously or sequentially, as required. Each camera supports a maximum video resolution of $1920 \times 1080$ pixels and a still-image resolution of $4608 \times 2592$ pixels. Video duration, frame rate, and time-lapse interval are user-configurable through the control interface.

The cameras are mounted on adjustable flexible brackets attached to the AGRICAM chassis, enabling field of view adjustment. Image data are transmitted to the onboard processing unit via HDMI cables.

In addition to imaging, an integrated environmental sensing module measures ambient temperature, relative humidity, and barometric pressure. The sensor is positioned adjacent to an air vent and is thermally isolated from electronic components to minimise interference. These sensors' data enable analysis of the relationship between pollinator activity and microclimate.


\subsection{Implementation and field evaluation}

\begin{figure*}[h!] 
    \centering 

    \begin{subfigure}[t]{0.98\textwidth}
        \centering
        \includegraphics[width=\textwidth]{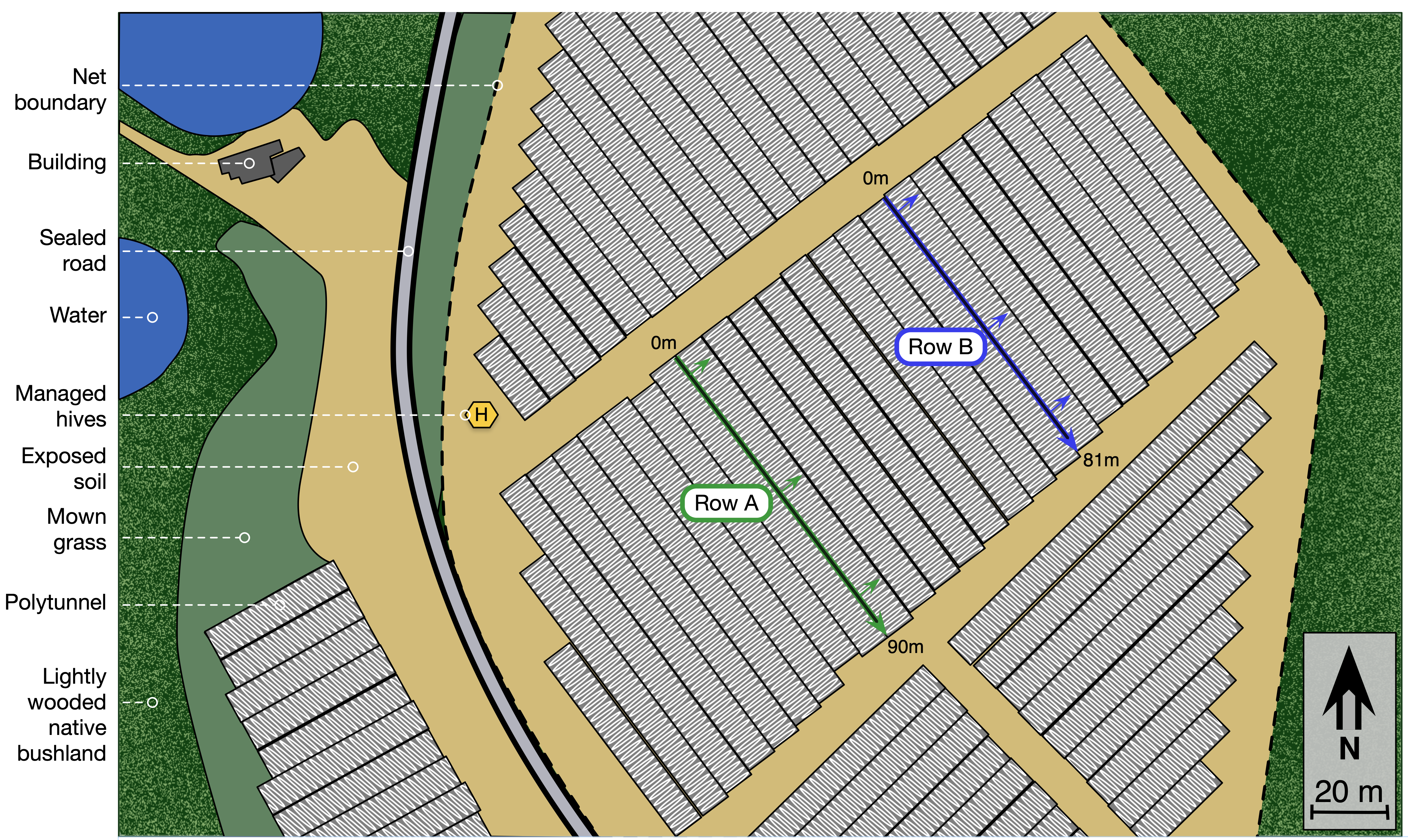}
        \caption{}
        \label{fig:costa_map}
    \end{subfigure}
    
    \vspace{1em} 

    \begin{subfigure}[t]{0.47\textwidth}
        \centering
        \includegraphics[width=\textwidth]{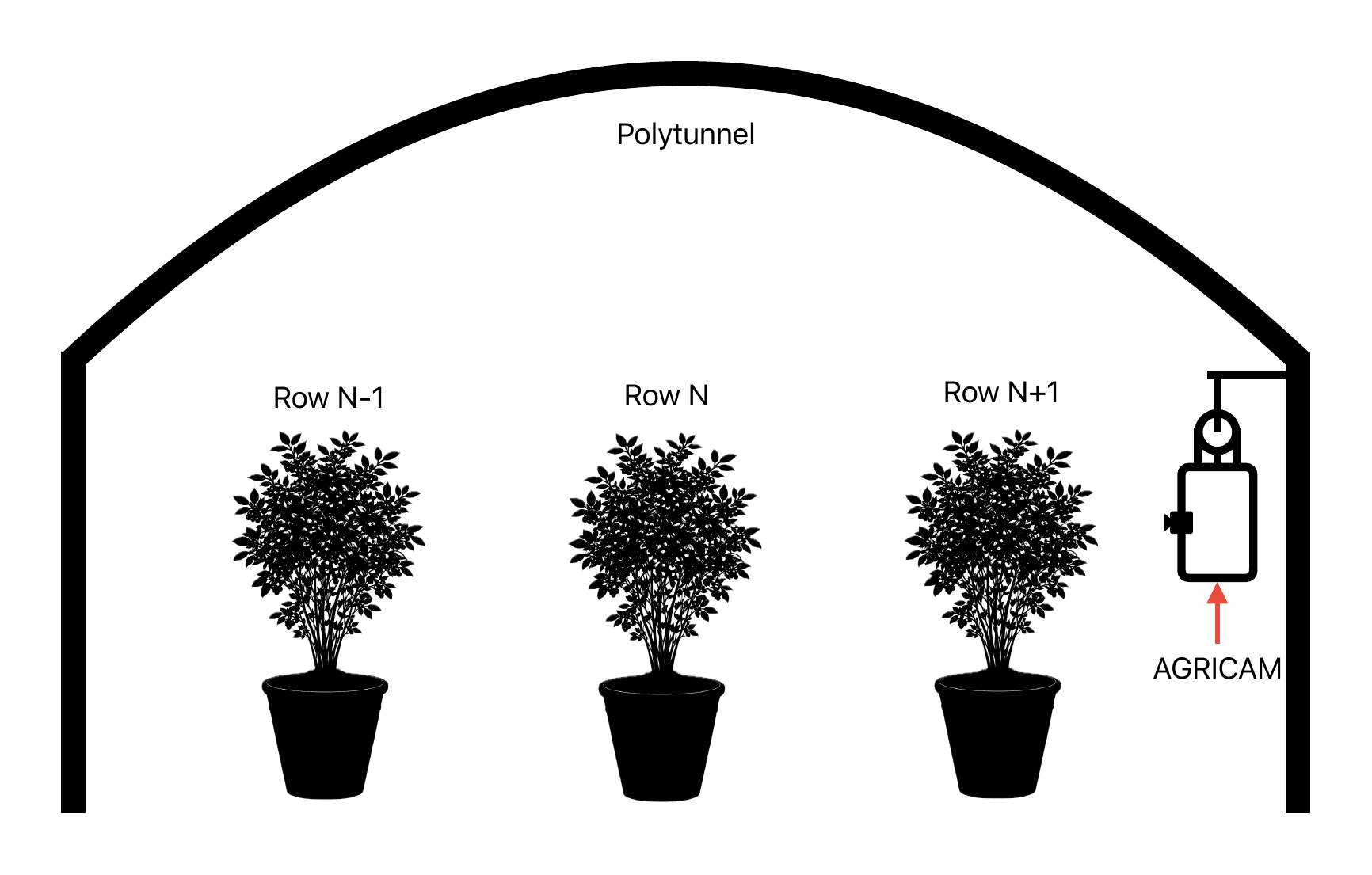}
        \caption{}
        \label{fig:tunnel_configuration}
    \end{subfigure}
    \hfill
    \begin{subfigure}[t]{0.47\textwidth}
        \centering
        \includegraphics[width=\textwidth]{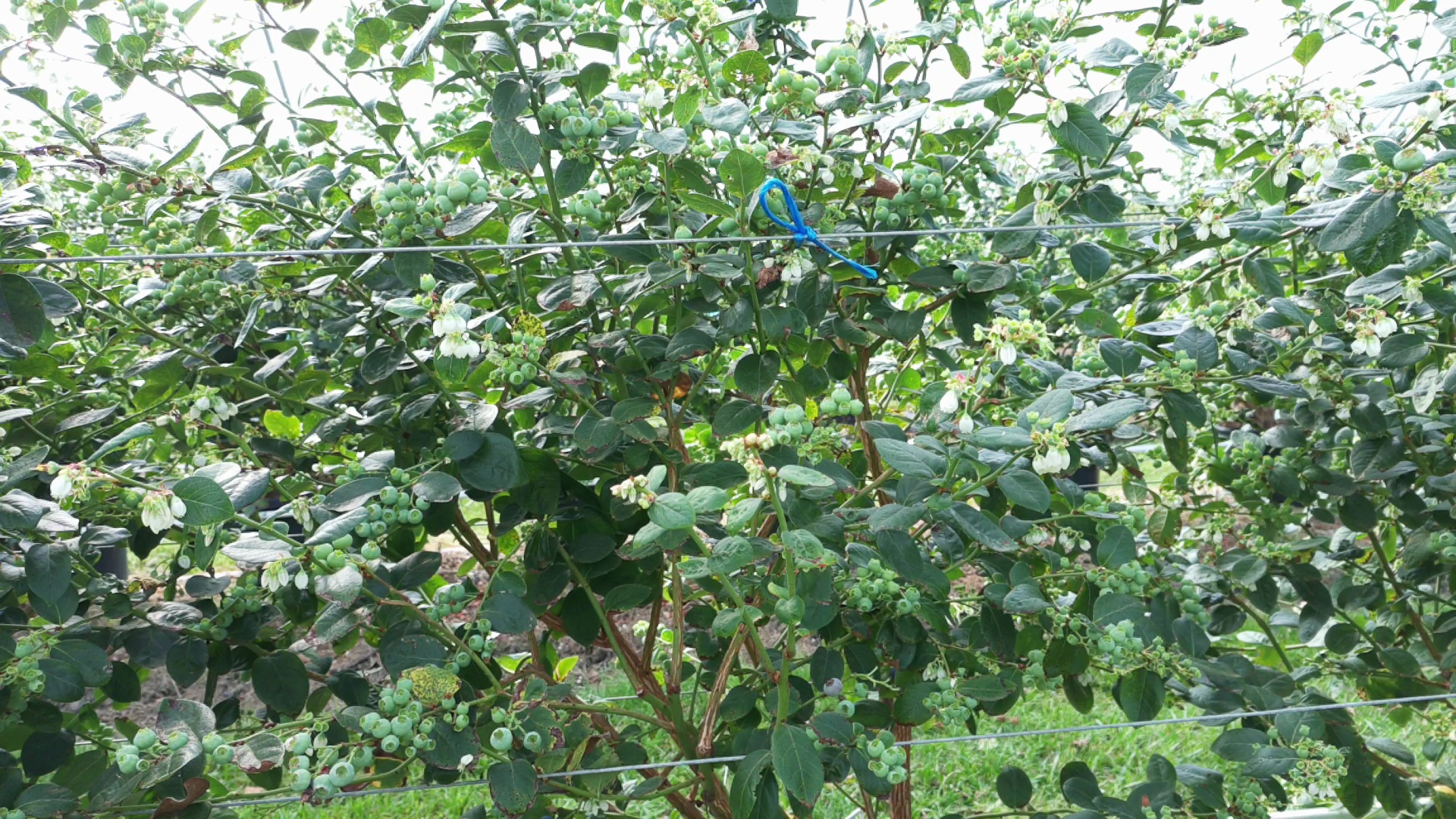}
        \caption{}
        \label{fig:costa-screenshot}
    \end{subfigure}

    \caption{\textbf{AGRICAM field test site detail.} (a) A map of the study site at Costa Berries Exchange, New South Wales, Australia. The location of managed beehives is indicated by a yellow pentagon. The crop rows where AGRICAM was deployed are labelled Row~A and Row~B, and the length of each monitored tunnel is indicated. Arrows show the direction the AGRICAM camera faces during row traversal. (b) Schematic representation of the tunnel configuration and blueberry row numbering, shown as viewed from the polytunnel end marked 0~m in (a). (c) Example image illustrating the field of view captured by AGRICAM.}

    \label{fig:implementaton}
\end{figure*}

We deployed AGRICAM to monitor insect pollination within commercial blueberry polytunnels at Costa Berries Exchange, New South Wales, Australia (Fig.~\ref{fig:costa_map}). The study was conducted under standard commercial production conditions towards the end of the peak flowering stage on 29--30 July 2025. The study site comprised multiple polytunnels, each containing three potted blueberry plant rows (Fig.~\ref{fig:tunnel_configuration}) where we installed two independent rail tracks in two separate polytunnels, A and B (Fig.~\ref{fig:costa_map}). Row~A and B were approximately 90~m and 81~m long respectively (Fig.~\ref{fig:costa_map}). Both tunnels were covered with translucent low-density polyethylene (LDPE) diffusing plastic and then within white bird netting (20 mm hole size) chosen by the growers.

The nearest managed honeybee hives were located 55~m from Row~A and 120~m from Row~B. To enable pollination monitoring under typical operating conditions, hive placement and crop management were not modified from standard farm procedure for the purpose of this study.

\subsubsection{Track installation and data collection}

The AGRICAM track was constructed using 6~m sections of standard 40~mm pressure-rated PVC pipe. Adjacent sections were joined using 120~mm-long inserts of 40~mm non-pressure PVC pipe to create flush connections for smooth robot traversal. Limit-switch triggers were installed at track ends (Figure~\ref{fig:system_installation_c}). 

AGRICAM monitoring stations were positioned along each plant row at intervals of 2.25--2.75~m for a total of 41 and 36 stopping stations on Rows~A and~B, respectively. Of these, AGRICAM was programmed to collect data at all station excluding terminal stations. AGRICAM was configured to initiate a new monitoring cycle at 15~min intervals throughout the daily operational period from 09:00 to 17:00, corresponding to the main activity window at the study site. At each monitoring station during track traversals out and back, microclimatic data was recorded at 15~s intervals, and 30~s video clips were captured using a single onboard camera. A 256 GB SD card was used for onboard data storage. After deployment, recorded data were transferred to external storage via a local area network (LAN) for offline processing.

\subsubsection{Video processing}
\label{sec:video_processing}

Insect tracking software \textit{Polytrack}~\citep{ratnayake2023spatial} was used to extract motion trajectories of insect pollinators and to detect blueberry flowers from the recorded videos. Polytrack employs a hybrid detection framework~\citep{ratnayake2021tracking} that combines foreground--background segmentation with deep learning–based object detection to identify and track insects and to record flower visits. The framework requires two separately trained object detection models, one for insect pollinators and one for flowers. We trained two YOLOv12~\citep{yolo12, tian2025yolo12} models for these tasks.

For insect detection, we developed a custom dataset of 2,844 annotated images containing honeybees, with a total of 3,271 labelled instances. The dataset was created and managed using Roboflow~\citep{dwyer2026roboflow}. We focused exclusively on honeybees, as they are the managed pollinator species at the study site and the only pollinator observed at the tunnels during the study period. Each image contained one or more honeybee instances annotated with bounding boxes using Roboflow. During annotation, we included only instances where the insect body was clearly visible to ensure high-quality ground truth labels. The dataset was split into training, validation and test sets in a 70/20/10 ratio. Horizontal and vertical flips were applied as data augmentation during training. We trained the YOLOv12 insect detection model from scratch using the Ultralytics framework. Training was conducted on an NVIDIA A40 GPU (MASSIVE M3)~\citep{goscinski2014multi} with a learning rate of 0.001 for 250 epochs. On the test split, the model achieved a recall of 97.5\% at a confidence threshold of 0.001. Based on this performance, we set the insect detection confidence threshold to 0.001 for Polytrack inference to prioritise recall. For flower detection, we created a separate custom dataset of 100 annotated images of blueberry flowers. This was annotated using Roboflow~\citep{dwyer2026roboflow} and used to train a dedicated YOLOv12 model following the same training framework. During inference, we applied a confidence threshold of 0.10 for flower detection.

Within Polytrack, we implemented a Mixture of Gaussians (MOG2) algorithm~\citep{zivkovic2006efficient} as the foreground--background segmentation module to detect moving objects. We tuned the MOG2 parameters using a manually curated representative subset of videos collected under varying environmental and lighting conditions to improve robustness to plant movement and scene variability.

We processed all recorded videos offline on the MASSIVE high-performance computing infrastructure using an NVIDIA A40 GPU. 

\subsubsection{Insect trajectory dataset preparation}

We manually reviewed all insect trajectories extracted using Polytrack to remove false-positive tracks caused by background motion, including leaf movement, lighting fluctuations, and minor oscillations of the \textit{AGRICAM} hardware due to wind-induced polytunnel movement. We retained only trajectories that visually corresponded to true insect motion for subsequent analysis.

For each validated trajectory, we represented the insect position as
\begin{equation}
P_i = (n_i, x_i, y_i),
\end{equation}
where $n_i$ denotes the video frame index and $(x_i,y_i)$ the corresponding two-dimensional image coordinates. We truncated each trajectory at the last frame containing valid coordinates to remove trailing missing values.

To fill missing detections within a trajectory, we applied linear interpolation along the temporal axis. For consecutive valid detections at frames $n_m$ and $n_{m+1}$, we estimated intermediate coordinates for frames $n_i \in [n_m,n_{m+1}]$ as
\begin{equation}
\alpha_i = \frac{n_i - n_m}{n_{m+1} - n_m},
\end{equation}
\begin{equation}
x_i^{\mathrm{interp}} = (1-\alpha_i)x_m + \alpha_i x_{m+1},
\end{equation}
\begin{equation}
y_i^{\mathrm{interp}} = (1-\alpha_i)y_m + \alpha_i y_{m+1}.
\end{equation}
This interpolation assumes approximately smooth motion between consecutive detections and reduces fragmentation caused by brief tracking losses.

We then smoothed the interpolated coordinates using a Savitzky--Golay (SG) filter~\citep{savitzky1964smoothing}, applied independently to the $x$ and $y$ sequences. The SG filter performs a local polynomial regression within a sliding window of length $w$ frames (default $w=11$) using a polynomial order of 2. When the trajectory length was shorter than the requested window, we set
\begin{equation}
w = \min(11, N),
\end{equation}
where $N$ is the number of samples in the trajectory, and enforced an odd-valued window length. If $w < 3$ or insufficient for polynomial fitting, we retained the interpolated coordinates without further smoothing. After filtering, we rounded the smoothed coordinates to the nearest integer pixel location:
\begin{equation}
x_i^{\mathrm{sm}} = \mathrm{round}\!\left(x_i^{\mathrm{SG}}\right),
\end{equation}
\begin{equation}
y_i^{\mathrm{sm}} = \mathrm{round}\!\left(y_i^{\mathrm{SG}}\right).
\end{equation}

We used insect flower visits as a proxy for pollination activity within the crop. To identify potential insect--flower visits, we associated each smoothed insect position $(s_x,s_y)$ with detected blueberry flowers using spatial proximity. For each video, we computed flower centre coordinates $(c_x,c_y)$ and flower radii $r$ (pixels). We computed the Euclidean distance between the insect and each flower centre as
\begin{equation}
d = \sqrt{(c_x - s_x)^2 + (c_y - s_y)^2}.
\end{equation}
We consider an insect to be close enough to a flower for a potential visit when
\begin{equation}
d \le r \cdot \eta,
\end{equation}
where $\eta$ is a radius extension factor (default $\eta=2.5$) introduced to account for localisation uncertainty and partial occlusion. When multiple flowers satisfied this condition, we assigned the insect to the nearest flower (minimum $d$). If no flower satisfied the criterion, we labelled the frame as unassigned.

We defined a flower visit based on the temporal continuity of flower proximity assignment. Let $F_i$ denote the proximate flower identifier associated with frame $i$, and let $T$ represent the minimum number of consecutive frames required to define a visit. A flower visit was recorded when an insect remained assigned to the same proximate flower for $T$ or more consecutive frames:
\begin{equation}
F_i = F_{i+1} = \dots = F_{i+k}, \quad k+1 \ge T.
\end{equation}

The threshold value $T$ was determined through manual review of a subset of videos containing 10 confirmed flower visits. In this sample, insects required a minimum of 10 frames ($\bar{x}$ = 20 frames, $s$ = 10 frames) to land on a flower after entering the predefined flower-radius region. To maximise recall and avoid excluding genuine visits, we selected $T = 10$ frames as the operational threshold. If an insect departed from a flower and subsequently re-entered the same flower after an absence of at least five consecutive frames, the re-entry was classified as a separate visit.

To reduce short gaps caused by single-frame missed detections, we applied a conservative temporal filling rule. When
\begin{equation}
F_{i-1} = F_{i+1} \neq \varnothing \quad \text{and} \quad F_i = \varnothing,
\end{equation}
we reassigned the intermediate frame to the same flower.

We used the resulting smoothed trajectories and visit classifications in all subsequent analyses, including flower-visit detection and quantification of pollinator activity.

\section{Results}
\label{sec:results}

\subsection{System deployment evaluation}

\begin{figure*}[h]
    \centering
    \includegraphics[width=\textwidth]{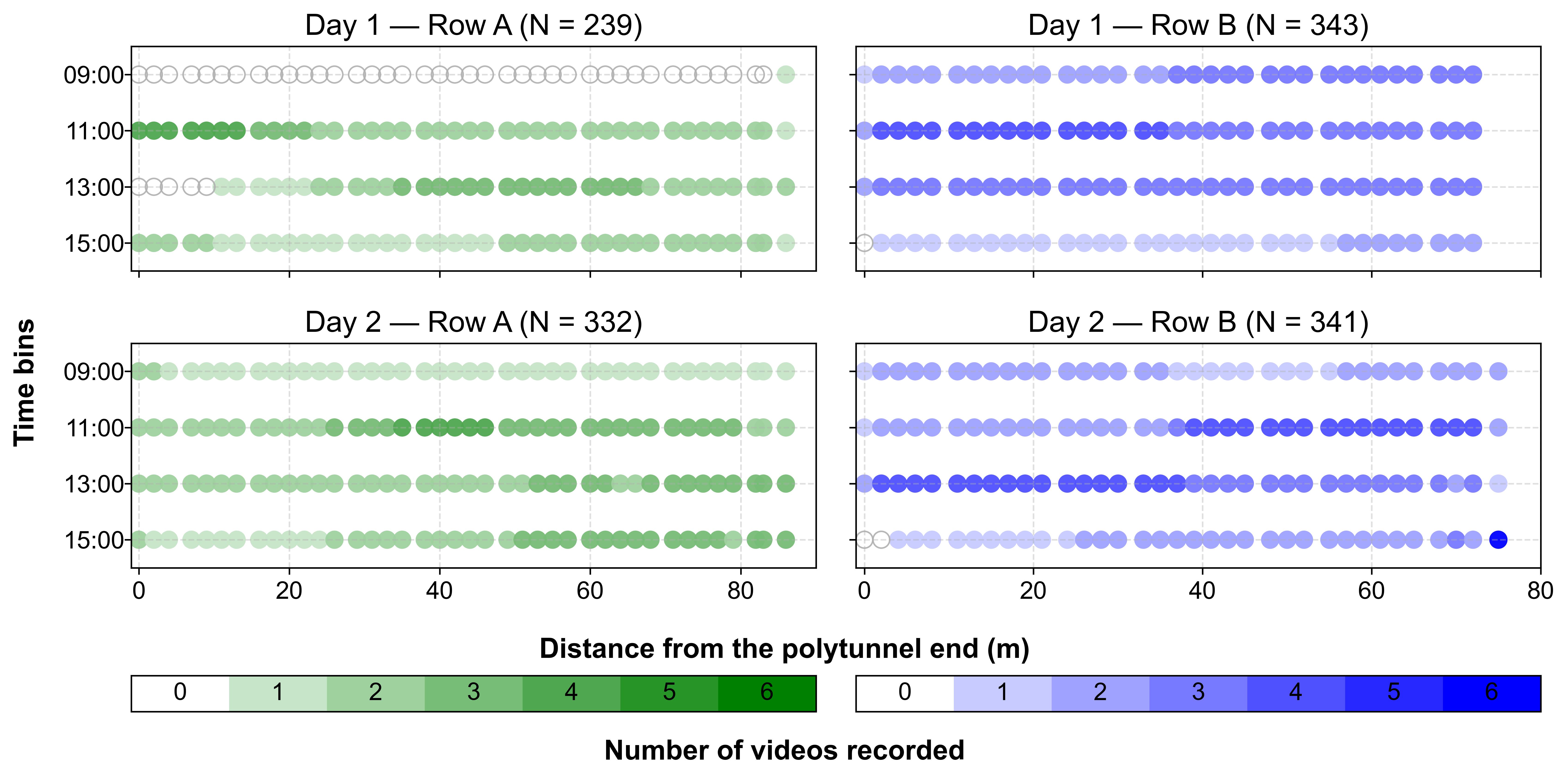} 
    \caption{\textbf{Spatial and temporal distribution of video recordings during field deployment.} Bubble plots show the number of videos recorded (colour scale 0--6) by distance from the polytunnel end (meters, x-axis) and by 2-hour time bin (09:00–, 11:00–, 13:00–, 15:00–16:59; y-axis) for two crop rows (Row~A - green, Row~B - blue) across two days. Panel titles report the total number of observations ($N$) for each day and row.}
    \label{fig:video_distribution}
\end{figure*}

We deployed AGRICAM under field conditions to evaluate its performance for automated insect pollination monitoring. We collected video footage and spatio-temporally distributed temperature and relative humidity data over two consecutive days from two blueberry rows (Rows A and B) at the study site. After set-up, the system operated without human intervention and without disrupting routine farm operations throughout the day; the robot was removed each evening and reinstalled in the morning due to vandalism concerns. {Each daily monitoring window (09:00--17:00) was completed on a single battery charge, without an in-day recharge or a low charge return-to-base event.} Altogether, a total of 1,249 videos were recorded. The spatiotemporal distribution of these recordings is shown in Figure \ref{fig:video_distribution}.

AGRICAM recorded a comparable number of videos across monitoring rows and sampling days, except on Day~1, Row~A, 09:00--11:00 am whilst this robot was being installed. After this, video acquisition was consistent and reliable across rows and days under field conditions. On average, each monitoring point in Rows~A and ~B was observed for approximately 82 and 86~seconds during the peak activity periods (11:00--13:00 and 13:00--15:00, respectively). In contrast, mean monitoring duration during the early morning (09:00--11:00) and late afternoon (15:00--17:00) time slots was lower, averaging 51~seconds per monitoring point. Overall, this corresponded to a mean daily monitoring time of 251~seconds (approximately 4.18~minutes) per monitoring station per row. The reduced sampling intensity in the early morning and late afternoon reflects operational constraints including initial setup and early termination of monitoring in response to limited honeybee activity observed during these periods. This level of fine-scale spatiotemporal coverage would be impractical to achieve using manual observations and surpasses the resolution typically attainable with existing stationary digital vision or acoustic pollinator monitoring systems.

\subsection{Insect detection and flower visit performance validation}

To evaluate the performance of insect track and flower visit detections generated by \textit{Polytrack} using data from AGRICAM, we conducted a validation study using 12 video samples (approx. 10,800 frames), comprising three randomly selected videos from each monitored row on each sampling day. Consequently, the validation captured representative commercial operating conditions, including naturally varying illumination, wind conditions, and canopy structures. For validation, \textit{Polytrack}-extracted insect tracks were overlaid onto the original video footage and compared with human observations that were treated as ground truth. Overlay videos were reviewed at 25\% playback speed using VLC Media Player. Upon the appearance of an insect, its movement was tracked from first detection to disappearance to record the number of flowers visited. When multiple insects were present simultaneously, individuals were assessed separately from their first appearance to ensure accurate visit quantification.

Insects detected in fewer than five consecutive frames were excluded, as at 30 frames per second this duration ($<0.17$ seconds) is too brief for reliable validation and is unlikely to generate meaningful pollination~\citep{ratnayake2023spatial}. An insect that exited the frame and subsequently re-entered was classified as a new individual. Similarly, insects temporarily occluded by foliage and later reappearing were recorded as new individuals. Only insects foraging on the camera-facing side of the crop were included in the analysis. Individuals on the opposite side of the blueberry bush were excluded, as indicated by movement paths that were spatially separated from, or visually obstructed relative to, the defined camera-facing region of interest. Accurate detection of a honeybee and successful generation of a corresponding track were classified as \textit{True Positives (TP)}, whereas tracks generated from non-honeybee movements, or duplicate tracks generated for the same honeybee after temporary tracking loss, were classified as \textit{False Positives (FP)}. Missed honeybee detections were classified as \textit{False Negatives (FN)}. A manual flower visit observation was defined as an instance in which an insect was visually confirmed to land on the dorsal (flowering) region of the plant within the camera’s field of view. Results of the study are presented in Table~\ref{table:polytrack_validation}.

\begin{table*}[h!]
\centering
\caption{\textbf{Validation of automated insect and flower visit detection using the Polytrack software.} ``Row'' and ``Date'' indicate the video samples analysed, and ``No.\ of videos'' the number evaluated. Under insect detection, ``Human obs.'' represents manually counted honeybees, while ``TP'', ``FP'', and ``FN'' denote true positives, false positives (including duplicate tracks of a single insect), and false negatives, respectively. Under flower visit detection, ``Human obs.'' indicates manually recorded insect--flower interactions and ``Software obs.'' those identified by the software.}
\small
\begin{tblr}{
  width = \linewidth,
  colspec = {Q[69]Q[83]Q[98]Q[175]Q[46]Q[46]Q[48]Q[169]Q[190]},
  cells = {c},
  row{2} = {font=\bfseries},
  row{7} = {font=\bfseries},
  cell{1}{1} = {r=2}{font=\bfseries},
  cell{1}{2} = {r=2}{font=\bfseries},
  cell{1}{3} = {r=2}{},
  cell{1}{4} = {c=4}{0.314\linewidth,font=\bfseries},
  cell{1}{8} = {c=2}{0.359\linewidth,font=\bfseries},
  cell{7}{1} = {c=2}{0.152\linewidth},
  hline{1,3,7-8} = {-}{},
  hline{2} = {4-9}{},
}
Row   & Date  & {\textbf{No. of~}\\\textbf{videos }} & Insect detection &    &    &    & Flower visit detection &               \\
      &       &                                      & Human obs.~      & TP & FP & FN & Human obs.             & Software obs. \\
A    & Day 1 & 3                                    & 6                & 5  & 2  & 1  & 7                      & 3             \\
A    & Day 2 & 3                                    & 10               & 10 & 1  & 0  & 2                      & 2 \\
B    & Day 1 & 3                                    & 13               & 10 & 1  & 3  & 8                      & 8             \\
B    & Day 2 & 3                                    & 10               & 8  & 0  & 2  & 3                      & 8             \\            
Total &       & 12                                   & 39               & 33 & 4  & 6  & 20                     & 21            
\end{tblr}
\label{table:polytrack_validation}
\end{table*}

Across the 12 validation videos (Table~\ref{table:polytrack_validation}), a total of 39 honeybees were recorded by manual observation. The Polytrack software correctly detected 33 of these individuals (true positives), with 6 false negatives and 4 false positives, resulting in an overall recall of 84.6\% and a precision of 89.2\%. Most false positives arose from duplicate tracks generated for a single honeybee, particularly when they remained stationary on a flower for extended periods or were partially occluded, leading Polytrack to reinitialise detections. Flower visitation counts were similar for manual and automated observations (20 vs.\ 21 visits in total), although there were differences in specific samples. For row B on Day 2, the software recorded more visits than were manually observed (8 vs.\ 3), whereas for row A on Day 1 it recorded fewer visits (3 vs.\ 7). Variation arose due to challenges in detecting individual blueberry flowers within dense cluster inflorescences using YOLO. In addition, flower movement due to wind, particularly when an insect was present, resulted in single visit being recorded as multiple interactions. Overall, the validation results indicate that Polytrack can quantify honeybee presence and flower visitation to a level of accuracy, that, as shown in the next section is suited to our application. However, as computer vision techniques continue to improve, AGRICAM will immediately benefit.

\subsection{Pollination analysis}

\subsubsection{Spatiotemporal patterns of honeybee activity}

\begin{figure*}[h]
    \centering
    \includegraphics[width=\textwidth]{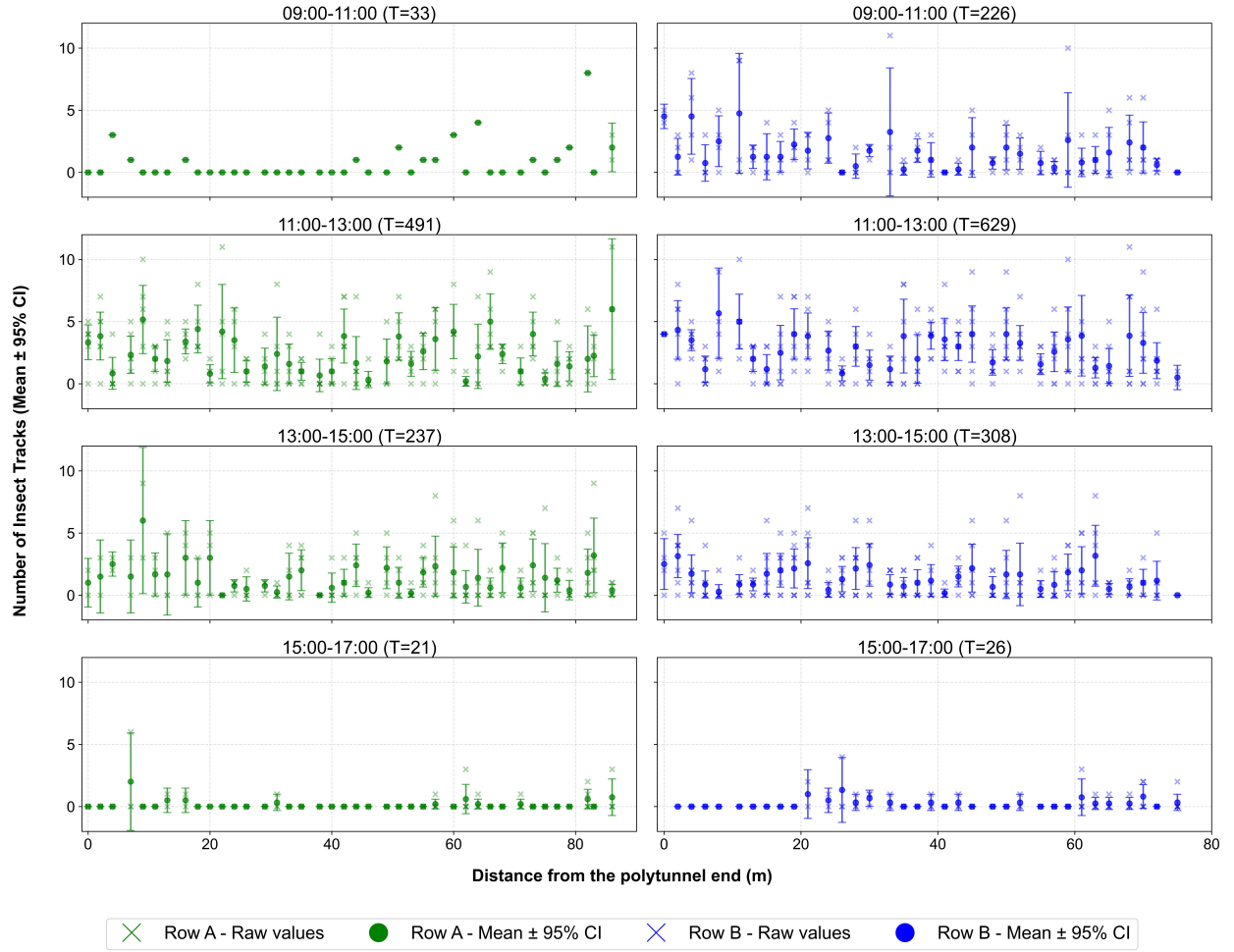} 
    \caption{\textbf{Spatiotemporal variation in insect activity across two blueberry rows.} Mean number of insect tracks (points $\pm$ 95\% CI) recorded along the length of Row A (green, left column) and Row B (blue, right column) during four two-hour time blocks (09:00--17:00) across two sampling days. X-axis = distance from polytunnel end (m), y-axis = number of tracks. Crosses (x) denote raw values. The time period, and total number of tracks per row ($T$), appear above each panel.}
    \label{fig:insect_tracks}
\end{figure*}

\begin{figure*}[h]
    \centering
    \includegraphics[width=\textwidth]{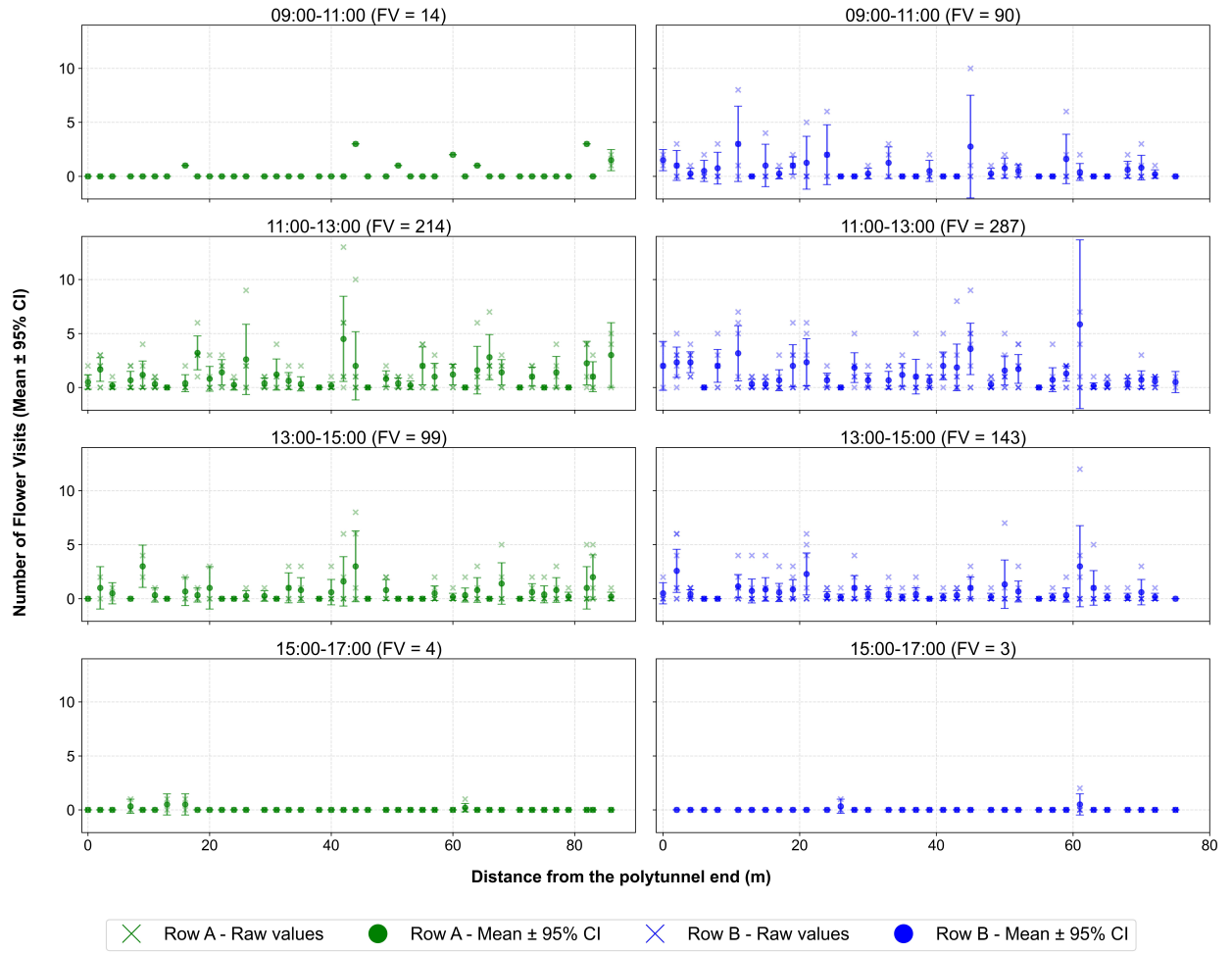} 
    \caption{\textbf{Spatiotemporal variation in flower visits across two blueberry rows.} Mean number of flower visits (points $\pm$ 95\% CI) recorded along the length of Row A (green, left column) and Row B (blue, right column) during four two-hour time blocks (09:00--17:00) across two sampling days. X-axis = distance from the polytunnel end (m), y-axis = number of flower visits. Crosses (x) denote raw values. The time period, and total number of flower visits per row ($FV$), appear above each panel.}
    \label{fig:flower_visits}
\end{figure*}

We analysed insect tracks extracted from AGRICAM-recorded videos to quantify the spatiotemporal variation in pollination activity across the two monitored blueberry rows. A total of 1,249 video sampling points were processed, yielding 1,971 honeybee tracks and 854 flower visits. These results provide a fine-scale characterisation of pollinator activity along rows by time-of-day. The results of the spatiotemporal analyses for insect tracks and flower visits are presented in Figures~\ref{fig:insect_tracks} and~\ref{fig:flower_visits}, respectively.

\subsubsection{Microclimatic drivers of honeybee activity}

\begin{figure*}[h]
    \centering
    \includegraphics[width=\textwidth]{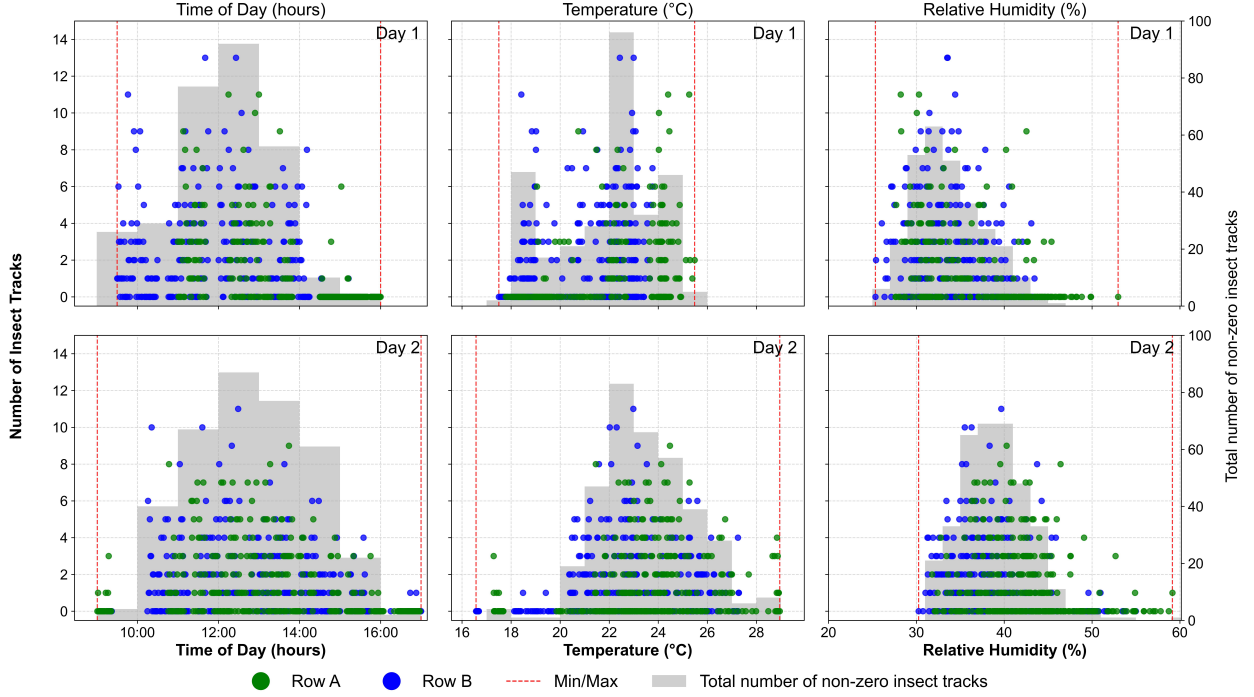} 
    \caption{\textbf{Distribution of insect tracks in relation to time of day and microclimate across two sampling days.} Number of insect tracks recorded in Row A (green) and Row B (blue) plotted against time of day (hours), temperature ($^\circ$C), and relative humidity (\%) for Day~1 (top row) and Day~2 (bottom row). Each point represents the number of bee tracks observed during an individual sampling interval. Grey histograms show the total frequency distribution of non-zero insect track observations across each environmental variable. Red dashed vertical lines indicate the minimum and maximum values recorded for each variable during the sampling period. The left y-axis shows the number of insect tracks per observation, and right y-axis (histograms) shows the total number of non-zero insect track observations.}
    \label{fig:insect_distributions}
\end{figure*}

We analysed honeybee activity in relation to time of day, temperature, and relative humidity to evaluate the influence of microclimate (Figure~\ref{fig:insect_distributions}). Temperature and relative humidity were recorded concurrently with insect activity at each monitoring location along polytunnels. Honeybee activity exhibited a diurnal pattern across the two monitoring days. Ambient temperature during the monitoring period ranged from 17$^\circ$C to 30$^\circ$C (Figure \ref{fig:insect_distributions}, Temperature). Peak honeybee activity was associated with temperatures between 22--23$^\circ$C. Relative humidity varied from 25\% to 60\%, with distinct daily patterns across the two sampling days (Figure \ref{fig:insect_distributions}, Relative Humidity). The highest insect counts generally coincided with intermediate humidity levels. Collectively, these results demonstrate how integrating fine-scale microclimate measurements with spatially distributed pollinator monitoring provides insight into their relationship (see Discussion).

\section{Discussion}
\label{sec:discussion}

\subsection{AGRICAM design and implementation}
\label{sec:disc:agricam}

In this study, we presented AGRICAM's application to automated pollinator monitoring in commercial blueberry polytunnels. We operated the system with minimal human intervention and without disrupting routine farm activities. This practical demonstration is important because adoption of agricultural robotics depends not only on technical performance, but also on integration with existing farm operations~\citep{lemay2024determinants}.

Most agricultural field robots operate on the ground using wheels or tracks \citep{bac2014harvesting, bechar2016agricultural}. In contrast, AGRICAM travels along an overhead track made from light, cheap, readily available components. The design has little structural impact within a polytunnel, and is easy to retrofit.  By removing ground contact, we avoided soil compaction and muddy-track creation, we didn't crush cover crops planted for beneficial insects, and we did not obstruct maintenance or harvesting pathways. Of direct benefit to the robot, our hardware was kept clear of dirt and farm debris and supported by a low-friction, adjustable platform. In protected cropping infrastructure, where space is limited and traffic flow is critical, this configuration provides an operational advantage. In addition, by maintaining a consistent camera path and positions, we improved the repeatability of data acquisition for longitudinal monitoring of flowering intensity and pollinator activity. In addition, our system minimises the need for human presence in polytunnels during monitoring. This is relevant because pollinator behaviour can be influenced by human movement, vibration and proximity \citep{delaplane2013standard}. AGRICAM also helps to standardise and reduce potential observer bias compared with manual field observations, while reducing labour costs for routine monitoring tasks.

A key limitation of conventional monitoring approaches is their reliance on stationary weather stations, which cannot capture the substantial spatial variability in temperature and relative humidity that occurs along polytunnels due to ventilation gradients, solar exposure, crop density, and structural orientation~\citep{hall2020bee, lei2023field, gruda2014protected, hou2021analysis}. By integrating microclimatic sensing with a mobile monitoring platform, our system addresses this limitation by capturing fine-scale environmental heterogeneity along the tunnel, providing measurements that more accurately reflect the conditions experienced by foraging bees.

By combining local RFID-based localisation with GNSS (GPS), our system is not impacted by weather and operates where satellite signals are weak inside polytunnels. Farmers can easily reposition RFID tags along a track to rapidly redefine monitoring zones to prioritise areas of agronomic interest without touching the robot.

Our single-motor drive design is simple, has low power requirements, and few potential points of mechanical failure. This ensures the system is reliable and easy to maintain.

A current limitation of the AGRICAM prototype is the requirement for manual battery recharging. Energy consumption depends on several operational factors, including polytunnel length and slope, monitoring duration and sampling frequency. AGRICAM's motor drive has been operated successfully on slopes up to 10 degrees, but in relatively flat polytunnels, motor load and battery consumption are reduced. Battery capacity presents a trade-off between operating time and system mass particularly relevant for mobile systems\citep{siegwart2011introduction} – larger batteries extend endurance but increase overall weight, which in turn increases rolling resistance and power demand. Future development should therefore consider optimisation of motor efficiency, lightweight structural materials and improved power management strategies. For commercial-scale deployment, an autonomous charging solution will be required. A docking station located at the end of the track could enable automatic recharging between monitoring cycles and overnight. Connection to the electricity grid or solar energy source would improve operational independence and reduce labour input.

\subsection{Automated video processing}


The Polytrack software~\citep{ratnayake2023spatial} used in this study was originally developed to track insects in two-dimensionally structured crops such as strawberry, where flowers are spatially separated and visible from above~\citep{ratnayake2021towards, ratnayake2023spatial}. Blueberry presents a more complex three-dimensional canopy structure of dense foliage and clustered flowers through which insects move and may be temporarily occluded from the camera. Current computer vision insect tracking approaches are limited in their ability to maintain insect identity through occlusions. As a result, single insect trajectories are often fragmented into multiple shorter tracks and the accuracy of visitation counts or behavioural metrics can be compromised. {Similar challenges occur during manual observations, where temporary occlusions may also prevent observers from reliably maintaining insect identity. Consequently, both manual and automated monitoring approaches are subject to uncertainty in visitation estimates under dense canopy conditions.} Future research should develop algorithms robust to this issue. Approaches such as multi-object tracking with temporal re-identification~\citep{zhang2021online}, motion modelling, or individual bee identification (e.g. appearance-based or marker-based identification)~\citep{zaman2024markerless} may improve trajectory continuity.

Although suspending AGRICAM's track from the polytunnel legs is cheap and simple, in strong winds we found some polytunnel structures sway, resulting in camera motion. This effect is more pronounced in shorter tunnels. Camera motion, combined with simultaneous movement of crops, flowers and insects, significantly increases scene variability, creating difficulties for tracking algorithms that may then generate false positive insect detections. Rapid plant movement can also obscure small pollinators, further complicating trajectory estimation. To address this limitation, future system development should consider both hardware and software stabilisation. Hardware-based solutions may include mechanical damping or lens stabilisation~\citep{xu2025video}. In parallel, software-based video stabilisation could reduce global camera motion before insect detection and tracking algorithms are applied. Combining mechanical stabilisation with digital correction is likely to provide the most robust solution.


Flower detection presents an additional challenge in blueberry crops. Unlike strawberry, where flowers are often spatially separated and individually distinguishable, blueberry flowers overlap in dense clusters. Polytrack currently uses a YOLO-based object detector~\citep{yolo12,yolo12} to identify flowers and draw bounding boxes around them. While this is effective with clearly separated flowers, it is less accurate in clustered blueberry inflorescences. Bounding-box detection does not adequately distinguish overlapping flowers, making it difficult to estimate the true number of flowers per cluster or the number of visits to individual flowers. Accurate flower-level visitation data would improve pollination assessment and could support more precise yield estimation. Future algorithm development should therefore explore instance segmentation or semantic segmentation approaches~\citep{yang2024detection, dias2018multispecies}. These methods can delineate individual flower boundaries within dense clusters, allowing more accurate counting and visit attribution. Improved flower segmentation would also enable extension of the monitoring framework to fruit detection and fruit classification based on size, colour and phenotype~\citep{maceachern2023detection}. Such integration would strengthen the link between pollination monitoring and yield prediction.

Our results highlight that automated pollination monitoring in three-dimensional bush crops presents challenges that are not fully addressed by systems developed for planar crops. Occlusion handling, motion robustness and accurate flower segmentation are critical areas requiring further investigation. By combining an overhead robotic monitoring platform with improved computer vision algorithms tailored to bush crop architecture, future systems may provide more accurate and scalable pollination assessment tools for protected horticulture.

\subsection{Pollination analysis}

Our results demonstrate that the proposed pollination monitoring system is capable of quantifying pollination activity in blueberry crops with high spatiotemporal resolution. Insect tracks (Figure~\ref{fig:insect_tracks}) and flower visits (Figure~\ref{fig:flower_visits}) were used as proxies for pollination in spatial intervals of 2.25--2.5~m along the polytunnels. To our knowledge, this level of detail  has not previously been achieved using either digital vision or acoustic monitoring tools. It would be impractical to obtain this amount of data through manual transect walks or quadrat-based observations.

Both insect tracks and flower visits were relatively uniform along the length of the monitored polytunnels (Figures~\ref{fig:insect_tracks} and~\ref{fig:flower_visits}). This contrasts with a previous study in blueberry and raspberry that reported reduced pollination activity in the middle of polytunnels, potentially attributed to microclimate~\citep{hall2020bee}. However, this study differed from the present work in several important respects, including the time of year when experiments were conducted, geographic location, and the length and configuration of the polytunnels. In addition, they relied on manual transect observations, which differed in sampling design and spatial coverage from our automated, continuous monitoring approach. Therefore, the difference in results may be attributed to site-specifc and/or methodological differences. Extended monitoring throughout the growing season would be required to further investigate the potentially variable distributions of insects on our study site. Such a study conducted with the spatial resolution our system provides, has the potential to detect fine-scale variation in pollinator abundance and to diagnose potential pollination deficits within specific tunnel sections at specific times of the year.

Pollinator activity exhibited a clear temporal pattern, with honeybee tracks per row increasing rapidly to a peak during the 2-hour period between 11:00 and 13:00, followed by a decline through the afternoon (Figure~\ref{fig:insect_tracks}; bracketed values (T = No.) and Figure~\ref{fig:insect_distributions}, time-of-day plots). The mean number of honeybee tracks showed a similar range between the two rows, despite differences in their distance from the nearest beehives (Figure~\ref{fig:insect_tracks}). The end of Row~A was located 55~m from the closest hives, whereas Row~B was 120~m away (Figure~\ref{fig:costa_map}), suggesting that, within the spatial scale of this study, distance from hive location to tunnel did not reduce foraging activity in the monitored rows. Flower visits, used here as a proxy for pollination activity, followed similar spatial and temporal patterns to insect track counts (Figure~\ref{fig:flower_visits}). However, the number of recorded visits was consistently lower than the total number of honeybee tracks across all time blocks, as not all detected bees engaged in clearly observable flower interactions within the camera’s field of view. Consequently, visit counts likely underestimate true pollination activity, as some interactions may be obscured by foliage.

\section{Conclusion}
\label{sec:conclusion}


In this paper, we presented AGRICAM, a {novel purpose-built robotic monitoring platform} for automated insect pollination monitoring in agricultural environments. Our system integrated a track-based mobile platform with vision-based sensing, RFID-based localisation, microclimate sensors, 4G connectivity, and a web-based interface to enable real-time data acquisition and control. {The data were subsequently analysed using} a computer vision and AI pipeline to perform automated, high-resolution spatio-temporal analysis of pollinator activity. We deployed and evaluated AGRICAM in 80~m long industrial polytunnels on a commercial farm over multiple days and demonstrated that our system captured spatial and temporal variations in insect activity, including their relationships with microclimatic conditions, while requiring minimal human intervention and operating without disrupting routine farm activities. {Field validation demonstrated scalable, high-resolution pollination monitoring under commercial conditions.} The presented system provides a scalable and practical solution for automated pollination monitoring. By generating rich, continuous data on insect--flower interactions, we enable more informed, data-driven pollination management. Our approach supports precision agriculture practices {through improved pollination management and informed decision-making}, contributing to improved food security at commercial scale.

\section*{CRediT authorship contribution statement}

\noindent Conceptualisation: MNR, ANT, JC, RR, AD; Data curation: MNR; Formal analysis: MNR, ANT, JC, RR, AD; Funding acquisition: ANT, JC, RR, AD; Investigation: MNR, ANT, JC, RR, AD; Methodology: MNR, ANT, AD; Project administration: MNR, AD; Resources: MNR, AD; Software: MNR; Supervision: AD; Validation: MNR, AD; Writing – original draft: MNR; Writing – review \& editing: MNR, ANT, JC, RR, AD.

\section*{Declaration of competing interest}

\noindent Alan Dorin, Adel N. Toosi, James Cook, and Romina Rader report financial support provided by the Australian Research Council. Romina Rader also reports financial support provided by the University of New England, Australia. The other authors declare that they have no known competing financial interests or personal relationships that could have appeared to influence the work reported in this paper.

\section*{Funding}
\noindent This research was supported by the Australian Government through the Australian Research Council Linkage Projects funding scheme (project LP210200213).

\section*{Data availability}
\noindent The datasets generated and/or analysed during this study are available from the corresponding author.

\section*{Acknowledgments}

\noindent The authors thank Mr. Maurizio Rocchetti, Mr. Brad Hocking and the Costa Berry Exchange, New South Wales, Australia staff for their assistance with track installation for the AGRICAM system. We also thank Mr. Jim Smiley for his assistance with the development of the AGRICAM robot. We are grateful to Mr. Harry Mills for his support with data annotation.

\section*{Declaration of generative AI and AI-assisted technologies in the manuscript preparation process.}

\noindent During the preparation of this work, the author(s) used ChatGPT and Codex (OpenAI) for code generation for data analysis. Following their use, the author(s) reviewed and edited the content as necessary and take full responsibility for the content of the published article.









\end{document}